\pdfoutput=1

\documentclass[11pt]{article}

\usepackage{ACL2023}

\usepackage{times}
\usepackage{latexsym}

\usepackage[T1]{fontenc}

\usepackage[utf8]{inputenc}

\usepackage{microtype}

\usepackage{inconsolata}
\usepackage{ulem}

\title{On the importance of Data Scale in Pretraining Arabic Language Models}

\author{Abbas Ghaddar$^{1\spadesuit}$\hspace{2mm} Philippe Langlais$^2$\hspace{2mm} 
\textbf{Mehdi Rezagholizadeh$^1$\hspace{2mm} Boxing Chen$^1$}\\
$^1$ Huawei Noah's Ark Lab\\
$^2$ RALI/DIRO, Universit\'e de Montr\'eal, Canada\\
\texttt{abbas.ghaddar@huawei.com}
}

\usepackage{amssymb}
\usepackage{pifont}

\newcommand{\bert}{BERT}
\newcommand{\arabert}{AraBERT}
\newcommand{\arbert}{ARBERT}
\newcommand{\marbert}{MARBERT}
\newcommand{\jaber}{JABER}

\newcommand{\saber}{SABER}

\newcommand{\ats}{AT5S}
\newcommand{\atb}{AT5B}
\newcommand{\artb}{AraT5-base}

\newcommand{\aramus}{AraMUS}

\newcommand{\jaberv}{JABERv2}
\newcommand{\jaberl}{JABERv2-6L}
\newcommand{\atsv}{AT5Sv2}
\newcommand{\atbv}{AT5Bv2}
\newcommand{\bom}{BO16*}

\newcommand{\mightmention}[1]{}
\newcommand{\problem}[1]{\textcolor{red}{$\star$}}
\newcommand{\answer}[1]{\textcolor{blue}{$\#$}}
\newcommand{\todoreview}[1]{\textcolor{green}{$@$}}

\usepackage{subcaption}
\usepackage{rotating,csquotes}
\usepackage{xcolor}
\usepackage{multirow}
\usepackage{framed}
\usepackage{booktabs}
\usepackage{amsmath}
\usepackage{upgreek}
\usepackage{ulem}
\usepackage{amsfonts}

\usepackage[most]{tcolorbox}

\newtcbox{\mybox}[1][]{enhanced, colframe=blue, colback=blue!15, 
	frame style={opacity=0.25}, interior style={opacity=0.25}, 
	nobeforeafter, tcbox raise base, shrink tight, extrude by=1mm, #1}

\newcommand\blfootnote[1]{%
  \begingroup
  \renewcommand\thefootnote{}\footnote{#1}%
  \addtocounter{footnote}{-1}%
  \endgroup
}

\begin{document}
\maketitle

\begin{abstract}

Pretraining monolingual language models have been proven to be vital for performance in Arabic Natural Language Processing (NLP) tasks. In this paper, we conduct a comprehensive study on the role of data in Arabic Pretrained Language Models (PLMs). More precisely, we reassess the performance of a suite of state-of-the-art Arabic PLMs by retraining them on massive-scale, high-quality Arabic corpora. We have significantly improved the performance of the leading Arabic encoder-only \bert{-base} and encoder-decoder T5-base models on the ALUE and ORCA leaderboards, thereby reporting state-of-the-art results in their respective model categories. In addition, our analysis strongly suggests that pretraining data by far is the primary contributor to performance, surpassing other factors. Our models and source code are publicly available at~\url{https://github.com/huawei-noah/Pretrained-Language-Model/tree/master/JABER-PyTorch}.

\blfootnote{$^{\spadesuit}$Corresponding author.}

\end{abstract}

\section{Introduction}

Pretraining~\cite{dai2015semi,devlin2018bert} is the most expensive and critical step in developing modern Language Models (LM)~\cite{brown2020language,min2023recent}. Its success involves numerous decisive design choices, such as data type, size, and quality~\cite{liu2019roberta,rae2021scaling,hoffmann2022training}, model architecture~\cite{raffel2019exploring, wang2022language,chowdhery2023palm}, and training algorithms~\cite{adafactor2018,santurkar2018does}. As these factors vary in importance, understanding their contributions is essential especially when targeting languages of different families than English\footnote{the mainstream language in NLP academic research.} or dealing with limited computing resources.

Despite the existence of dozens of Arabic Pretrained LM~\cite{antoun2020arabert,abdul2021arbert,ghaddar2022revisiting,nagoudi2023jasmine} and the progress made to standardize their evaluation~\cite{seelawi2021alue,elmadany2023orca,nagoudi2023dolphin}, the primary factor driving performance on NLP tasks remains unclear. In this work, we dive deep into studying the influence of pretraining data (size and quality) for performances of Arabic PLMs. We do so by first pretraining multiple new Arabic \bert{} and T5 models of different sizes on a massive scale and highly filtered Arabic corpora, following the best pretraining practices of Arabic PLM from the literature. Then, we conduct a comprehensive evaluation study on two well-established Arabic NLP benchmarks, namely ALUE~\cite{seelawi2021alue}  and ORCA~\cite{elmadany2023orca}.

Among factors like model size, architecture, and pretraining data, we found that the latter is the most significant for performance.  For instance, we found that an Arabic T5-small model, despite being 3 times smaller, can perform on par with a T5-base model when pretrained on four times more data. Similarly, a 6-layer \bert{} model performs on par with a 12-layer \bert{} model when both are trained on datasets of roughly the same size but with the latter using lower quality data. More importantly, our findings suggest that scaling up pretraining data is more critical for generative encoder-decoder models compared to encoder-only models. For example, we found that a \bert{-base} model only improves by 0.4\% on the ORCA score compared to a 3.1\% improvement for a T5-base model when scaling up the data size by a factor of 4. 
\begin{table*}[ht] 

\centering
 \renewcommand{\arraystretch}{1.25}
\resizebox{1\textwidth}{!}{%
\begin{tabular}{l@{\hspace{0.75\tabcolsep}}l@{\hspace{0.75\tabcolsep}}|c@{\hspace{0.75\tabcolsep}}c@{\hspace{0.75\tabcolsep}}|c@{\hspace{0.75\tabcolsep}}c@{\hspace{0.75\tabcolsep}}c@{\hspace{0.75\tabcolsep}}c@{\hspace{0.75\tabcolsep}}c@{\hspace{0.75\tabcolsep}}c@{\hspace{0.75\tabcolsep}}c@{\hspace{0.75\tabcolsep}}c}  
 \toprule  
\textbf{\colorbox{blue!0}{Cluster}}  &
\textbf{\colorbox{blue!0} {Task}}&
\textbf{\colorbox{purple!20}{{\arbert{v2}}}}  &
\textbf{\colorbox{purple!20}{{\bom}}}&
\textbf{\colorbox{yellow!30}{{\saber}}} &
\textbf{\colorbox{purple!20}{{\jaberv$^\dagger$  }}}&
\textbf{\colorbox{green!20}{{\atbv$^\dagger$ }}} &
\textbf{\colorbox{purple!20}{{\jaber }}} &
\textbf{\colorbox{gray!20}{{\jaberl$^\dagger$ }}}  & 
\textbf{\colorbox{blue!20}{{\atsv$^\dagger$  }}} &
\textbf{\colorbox{green!20}{{\atb }}} &
\textbf{\colorbox{blue!20}{{\ats }}}  
\\
\toprule 

\multicolumn{1}{l}{\multirow{11}{*}{\textbf{SC}}} 
& abusive  &  $76.0$  &  $79.7$  & $\textbf{82.2}$  & $78.2$ & $76.6$ & $\underline{81.3}$ & $77.0$ & $72.2$ & $73.5$ & $73.3$ \\
& adult & $89.7$ & \underline{\textbf{91.0}} & $90.8$ & $90.3$ & $90.5$ & $90.3$ & $90.1$ & $89.4$ & $90.3$ & $89.7$ \\
& age  & $45.8$ & \underline{\textbf{47.3}} & $45.9$ & $46.9$ & $45.4$ & $44.0$ & $45.0$ & $44.2$ & $43.4$ & $42.4$ \\
&  claim &  $67.4$  &  $\underline{71.5}$  &  $\textbf{73.1}$ & $69.4$ & $69.0$ & $70.8$ & $69.2$ & $70.2$ & $64.8$ & $66.0$ \\
 & dangerous  &  $65.0$  &  $67.3$  & $\textbf{67.9}$ & \underline{$67.8$} & $66.5$ & $66.8$ & $63.7$ & $61.8$ & $59.4$ & $61.1$ \\
& dialect-B & $86.9$ & \underline{$87.9$} & $\textbf{88.4}$ & $87.4$ & \underline{$87.9$} & \underline{$87.9$} & $87.7$ & $87.7$ & $86.8$ & $87.2$ \\			
&  dialect-R  &  $65.2$  &  $69.2$  &  $\textbf{71.8}$ & \underline{$69.6$} & $66.3$ & $65.5$ & $65.4$ & $65.8$ & $65.6$ & $65.0$ \\
&  dialect-C  &  $35.7$  &  $36.5$  & $\textbf{40.7}$ & $38.0$ & $37.6$ & \underline{$38.1$} & $36.1$ & $32.2$ & $34.2$ & $31.2$ \\
&  emotion-cls & $64.8$ & $\textbf{70.8}$ & \underline{$70.1$} & $69.6$ & $65.7$ & $67.2$ & $64.0$ & $61.0$ & $62.9$ & $59.8$ \\
& emotion-reg  & $67.7$ & $74.3$ & $\textbf{77.3}$ & \underline{$77.0$} & $75.3$ & $71.5$ & $74.2$ & $69.6$ & $68.7$ & $62.5$ \\
& gender  & $63.2$ & \underline{$\textbf{67.6}$} & $65.6$ & $67.5$ & $67.4$ & $65.0$ & $66.3$ & $65.7$ & $63.3$ & $62.1$ \\

 &  hate   &  $82.3$  &  $85.3$  &  $85.8$ & $\underline{\textbf{86.8}}$ & $84.0$ & $83.4$ & $81.8$ & $78.1$ & $80.6$ & $82.5$ \\
  &  irony  &  $83.8$  &  \underline{$85.2$}  &  $\textbf{85.7}$ & $82.5$ & $83.9$ & $84.8$ & $83.8$ & $82.7$ & $82.2$ & $81.4$ \\
  &  offensive  &  $89.6$  &  $92.2$  &  $\textbf{93.0}$ & \underline{$92.3$} & $91.7$ & $89.7$ & $90.4$ & $89.8$ & $90.8$ & $88.0$ \\

  &  machine G.  &  $87.9$  &  \underline{$\textbf{90.7}$}  & $90.1$ & $88.8$ & $87.4$ & $89.3$ & $87.4$ & $86.5$ & $88.3$ & $86.6$ \\
  &  sarcasm &  $74.2$  &  \underline{$76.8$}  &  $\textbf{77.3}$ & $74.3$ & $74.8$ & $74.2$ & $72.8$ & $72.1$ & $76.5$ & $71.8$ \\

& sentiment  & $78.6$ & \underline{$\textbf{80.9}$} & $80.2$ & $80.2$ & $79.7$ & $80.2$ & $80.2$ & $77.9$ & $79.0$ & $77.8$ \\ 
\hline

\multicolumn{1}{l}{\multirow{3}{*}{\textbf{SP}}} 
 &  ner-anerc.   &  $90.8$  &  $90.8$  &  $91.2$ & $90.4$ & $90.6$ & $90.7$ & $90.6$ & $90.8$ & \underline{$\textbf{91.4}$} & $89.7$ \\
  &  ner-aqmar   &  $81.7$  &  $81.7$  &  $\textbf{83.7}$ & $81.2$ & \underline{$82.5$} & $81.5$ & $81.1$ & $77.4$ & $78.7$ & $79.8$ \\
  & pos-dia & $92.9$ & \underline{$94.7$}  &  $\textbf{94.8}$ & $93.8$ & $93.7$ & $93.4$ & $93.6$ & $93.3$ & $93.3$ & $92.3$ \\ 
   &   pos-xglue    &   $52.6$ & \underline{$\textbf{69.4}$} & $59.3$ & $59.9$ & $57.5$ & $57.5$ & $51.1$ & $52.9$ & $30.5$ & $25.8$ \\ 
   \midrule
  
 \multicolumn{1}{l}{\multirow{3}{*}{\textbf{NLI}}} 

 &  ans-st   &  $91.0$  &  $93.2$  &  $91.3$ & $90.8$ & \underline{$\textbf{93.3}$} & $92.3$ & $87.7$ & $84.2$ & $80.7$ & $85.0$ \\
 &  baly-st   &  $49.3$  &  \underline{$\textbf{51.2}$}  &  $50.1$ & $46.1$ & $46.7$ & $48.8$ & $42.2$ & $49.2$ & $44.0$ & $45.1$ \\ 
  &  xlni  &  $68.2$  &  $70.2$  &  $71.1$ & \underline{$\textbf{71.4}$} & $72.0$ & $71.3$ & $67.1$ & $64.8$ & $71.7$ & $65.7$ \\\hline 
  
\multirow{2}{*}{\textbf{STS}}  

  &  sts-r  &  $71.9$  &  $76.0$  & $\textbf{78.8}$ & \underline{$77.4$} & $75.8$ & $75.1$ & $71.9$ & $74.2$ & $72.9$ & $73.2$ \\
  &  sts-c &  $96.7$  &  $97.1$  &  $95.8$ & \underline{$\textbf{98.8}$} & $96.2$ & $96.2$ & $94.7$ & $95.1$ & $95.3$ & $94.8$ \\\hline 
  \textbf{TC} &  
  topic   &  $93.9$  &  \underline{$\textbf{94.6}$}  &  $94.1$ & $93.5$ & $94.2$ & $93.7$ & $93.6$ & $94.0$ & $94.4$ & $93.9$ \\\hline

 \multirow{1}{*}{\textbf{QA}}  
   &  arlue-qa  &  $61.5$  &  $61.6$  & $62.8$ & \underline{$\textbf{63.2}$} & $62.8$ & $62.8$ & $56.6$ & $54.7$ & $62.0$ & $54.7$ \\ \hline
 
\textbf{WSD} &
  ar-wsd   &  $71.0$ & \underline{$\textbf{76.7}$} & $76.5$ & $74.9$ & $75.0$ & $75.4$ & $72.8$ & $70.4$ & $74.3$ & $71.3$ \\ \hline
							
  \toprule  
  
\multicolumn{2}{l}{\textbf{ORCA score}}&  $74.0$ & - & $\textbf{77.1}$ & \underline{$76.1$} & $75.5$ & $75.5$ & $73.7$ & $72.7$ & $72.4$ & $71.0$ \\ 			
\bottomrule  

\end{tabular} }

\caption{\small{Performance of models on ORCA leaderboard (test sets). We report the results of 8 models we finetune and submit ourselves to the ORCA leaderboard, as well as the results of \arbert{v2}, which is the top-ranked model on ORCA leaderboard. \textbf{\bom} column shows the per-task best score among the 16 Arabic PLM baseline studied by \cite{elmadany-etal-2023-orca}. Bold entries describe the best results among all models, while underlined entries show the best results among \textit{base} models. $\dagger$ marks the new models we pretrained in this work. Model names are highlighted to indicate their configuration: \colorbox{purple!20}{\bert{-base}}, \colorbox{yellow!30}{\bert{-large}},
\colorbox{green!20}{T5-base}},
\colorbox{gray!15}{6 layer \bert{}}, and 
\colorbox{blue!30}{T5-small}. 
}

\label{tab:orca_test}

\end{table*} 

We hope that our findings help to guide the design choices aimed at developing the next generation of Arabic Large Language Models (LLMs). This is particularly relevant as recent studies~\cite{khondaker2023gptaraeval,nagoudi2023jasmine,huang2023acegpt} suggest that newly emergent LLMs~\cite{touvron2023llama,muennighoff2023crosslingual}, including proprietary models like ChatGPT~\cite{chatgpt2022}, perform poorly on Arabic NLP tasks. 


\section{Experiments}
\label{sec:Experiments}

\subsection{Pretraining Setting}
\label{sec:Pretraining}

The goal is to study the impact of significantly scaling up the pretraining data on Arabic encoder-only \bert~\cite{devlin2018bert} and encoder-decoder T5~\cite{raffel2019exploring} style models. We conduct pretraining experiments on top of 3 Arabic PLM models from \cite{ghaddar2022revisiting}: \jaber{}, \ats{}, and \atb{} that correspond to the \bert{-base}, T5-small, and T5-base models, respectively. We choose the models and configurations of~\cite{ghaddar2022revisiting} since they demonstrate superior performance on ALUE. Additionally, those models utilize the same pretraining data, tokenizer, and vocabulary, facilitating comparisons across configurations. More precisely, we pre-train new \textbf{\jaberv{}}, \textbf{\atsv{}}, \textbf{\atbv{}} models by scaling up their pretraining data from 128GB~\cite{ghaddar2022revisiting} to 512GB of Arabic text data derived from 90 shards of Common Crawl monthly shards. Furthermore, to study the impact of data scale on smaller encoder-only models, we pretrained a 6-layer \jaber{}. This model will henceforth be referred to as \textbf{\jaberl{}}. Pretraining data, models, and implementation details are provided in Appendix~\ref{sec:Pretraining Details}.

\subsection{Finetuning Setting}
\label{sec:Finetuning}

We evaluate all models on 2 well-established Arabic NLP benchmarks: ORCA~\cite{elmadany2023orca} and ALUE~\cite{seelawi2021alue},
which include 29 and 8 tasks respectively. Since ORCA is a relatively new benchmark that does not include the models from \cite{ghaddar2022revisiting} that we are comparing, we fine-tuned and submitted: \jaber{}, \ats{}, \atb{}, as well as \saber{}, to the leaderboard using their respective publicly available repositories. In total, we conducted more than $5200$ fine-tuning experiments, which included searches for optimal hyperparameters and random runs, across both benchmarks. See Appendix~\ref{sec:Fine-tuning Details} for a  description of the datasets, splits, implementation details, and evaluation metrics. Results for the development sets on ORCA and ALUE are presented respectively in Tables~\ref{tab:orca_dev} and \ref{tab:dev_alue} in Appendix~\ref{sec:Dev Set Results}.

\subsection{ORCA Results}
\label{sec:ORCA Results}

Table~\ref{tab:orca_test} shows the performances of 8 models that we finetune and submit to the ORCA leaderboard. We also report the results of two baselines derived from existing models on the ORCA leaderboard: \arbert{v2}~\cite{elmadany2023orca}, which was the top-ranked model on the leaderboard, and a configuration we refer to as \bom{} (best of 16): for the 16 Arabic PLMs evaluated by~\cite{elmadany2023orca}, we pick up the score of the best-performing model for each task.  

We observe that \saber{} has become the new state-of-the-art model on the leaderboard, achieving an ORCA score of 77.1. This achievement is notable as it is the largest (\bert{-large}) model submitted to the leaderboard to date compared to its counterparts (\bert{-base}). While \saber{} reports the best performances on 12 out of 27 tasks, it also significantly underperforms compared to the ensemble of ORCA Arabic \bert{-base} baselines (\bom), particularly on tasks involving Arabic dialect data (e.g., pos-dial, age). On these tasks, Arabic PLM that are pretrained solely on Arabic dialect and tweet data perform the best. This is expected, considering \saber{} was pretrained exclusively on MSA data. \jaberv{} ranks second overall and first among models of the \bert{-base} size, outperforming \arbert{v2} by an average of 2.1\%. In addition, \jaberv{} demonstrates competitive performance against the aggregate of 16 Arabic PLM (\bom{}), outperforming them in 10 out of 27 tasks.  Furthermore, it closely matches these models, underperforming by only 0.5\% or less on another 7 tasks.

The benefits of scaling up the pretraining data are noticeable from three key observations. First, although \jaberl{} is half the size of \arbert{v2}, it underperforms by only 0.3\% on average. This is thanks to \jaberl{} being pretrained on a dataset larger than the one used for \arbert{v2}. Second, we observe that data scaling is more beneficial to encoder-decoder models (\ats{} and \atb{}) than to encoder-only ones (\jaber). The ORCA scores show an improvement of 1.4\% and 3.1\% when comparing \ats{} with \atsv{}, and \atb{} with \atbv{}, respectively. This is in contrast to a smaller increase of 0.6\% observed between \jaber{} and \jaberv{}. It's also interesting to see \atsv{} slightly outperforms \atb{} while being 3 times smaller, which further supports our claim that pretraining data is more important than model size for Arabic PLM.

Furthermore, we observe that scaling up the pretraining data also helps in narrowing the performance gap between encoder-decoder and encoder-only models. It reduces the gap from 3.1\% between \jaber{} and \atb{} to just 0.6\% between \jaberv{} and \atbv{}. This observation is in line with findings on English T5~\cite{raffel2019exploring}, suggesting that encoder-decoder models require more pretraining data compared to encoder-only models.

\begin{table*}[!htp]
\small
\centering
\resizebox{\textwidth}{!}{
\begin{tabular}{l|c|cccccccc|c}
\toprule
\textbf{Model} & \textbf{\#Params} & \textbf{MQ2Q}& \textbf{MDD}& \textbf{SVREG}& \textbf{SEC}& \textbf{FID}& \textbf{OOLD}& \textbf{XNLI}& \textbf{OHSD} & \textbf{Avg.}\\
\midrule
\multicolumn{10}{c}{\textit{Small-scale Models}}\\
\midrule
\ats & 109M & 85.3 & \underline{61.9} & 60.1 & 25.3 & 84.1 & 88.5 & 67.5 & 77.6 & 68.8 \\
\atsv$^\dagger$ & 109M & 85.5 & 59.3 & \underline{67.1} & 26.9 & \underline{84.5} & 87.3 & 67.6 & 75.9 & 69.3 \\
\jaberl & 86M$^\dagger$ & \underline{91.5} & 60.9 & 66.2 & \underline{28.2} & 84.4 & \underline{90.2} & \underline{69.0} & \underline{78.7} & \underline{71.1} \\

\midrule
\multicolumn{10}{c}{\textit{T5-base Models}}\\
\midrule

\atb & 296M & 86.3 & 63.4 & 62.0 & 28.7 & 84.4 & 89.3 & 72.4 & 78.6 & 70.6 \\
\artb & 282M & \underline{91.3} & \underline{63.8} & 65.9 & 30.5 & 82.3 & 88.8 & 68.2 & 77.9 & 71.1 \\
\atbv$^\dagger$ & 296M & 86.6 & 62.3 & \underline{71.7} & \underline{34.5} & \underline{84.7} & \underline{92.2} & \underline{74.2} & \underline{82.0} & \underline{73.5} \\

\midrule
\multicolumn{10}{c}{\textit{\bert{-base} Models}}\\
\midrule
\marbert{} & 163M & 83.3 & 61.9 & 75.9 & \underline{36.0} & \underline{85.3} & 92.1 & 64.3 & 78.9 & 72.2 \\
\jaber & 135M & 93.1 & \underline{64.1} & 70.9 & 31.7 & \underline{85.3} & 91.4 & 73.4 & 79.6 & 73.7 \\ 
\jaberv$^\dagger$ & 135M  & \underline{92.1} & 62.8 & \underline{72.1} & 34.7 & 84.4 & \underline{92.2} & \underline{73.6} & \underline{81.2} & \underline{74.1} \\

\midrule
\multicolumn{10}{c}{\textit{Large-scale Models}}\\
\midrule

ALM-1.0 & 350M & 94.5 & 65.1 & 70.1 & 35.3 & 86.0 & 91.7 & 77.7 & 85.7 & 75.8 \\
AceGPT & 13B  & 94.9 & 63.4 & 72.4 & 36.8 & 85.1 & 94.2 & 81.0 & 85.4 & 76.6\\
\saber  & 369M  & 93.3 & 66.5 & 79.2 & 38.8 &  86.5 & 93.4 & 76.3 & 84.1 & 77.3\\
\aramus{} & 11B & \bf \underline{95.2} & \bf \underline{67.5} &\bf \underline{80.4} & \bf \underline{41.6} & \bf \underline{87.2} & \bf \underline{95.5} & \bf \underline{83.2} & \bf \underline{87.4} & \bf \underline{79.8} \\
\bottomrule

\end{tabular}
}
\caption{ALUE Leaderboard test results of Arabic PLM as of 01/01/2024. The models are categorized into four groups based on their architecture and size. Entries in bold indicate the best results across all models, while entries that are underlined denote the best results within each respective category. $\dagger$ marks the new models we pretrained in this work.}
\label{tab:test_alue}
\end{table*}

\subsection{ALUE Results}
\label{sec:ALUE Results}

Table~\ref{tab:test_alue} presents the ALUE leaderboard test set results for a suite of Arabic PLM models, grouped according to their architecture and size. While many of the observations align with those noted on the ORCA benchmark, we observe that scaling up pretraining data has resulted in only a minor gain of 0.3\% on \ats{}. Additionally, there is a significant gap between \jaberl{} and a decent \bert{-base} baseline (\marbert{}). Both observations can be primarily attributed to the characteristics of the ALUE benchmark, where the tasks predominantly involve dialect data with a limited number of training examples. Besides, we also observe that the encoder-decoder model (\atbv) not only shows greater improvement with the scaling of pretraining data but also narrows the performance gap with the encoder-only model (\jaberv). 

It is interesting to note that, despite the significant differences in characteristics between the ALUE and ORCA benchmarks, we observe that the gap between models closely matches in terms of absolute values. For instance, the gap is 0.4\% and 0.6\% between \jaber{} and \jaberv{} on ALUE and ORCA respectively; 2.9\% and 3.1\% between \atb{} and \atbv{} on the same benchmarks; and exactly 0.6\% between \atbv{} and \jaberv{} on both benchmarks. The consistency of results across benchmarks gives more credibility to our conclusion regarding the importance of scaling data for Arabic LM pretraining.

Surprisingly, AceGPT~\cite{huang2023acegpt} not only underperforms compared to \aramus{}~\cite{alghamdi2023aramus}, an 11B Arabic T5 model pretrained on data comparable to the one we use, but also falls short against \saber{}, which is 35 times smaller. This is more surprising when knowing that AceGPT uses LLama 2~\cite{touvron2023llama} as its backbone and has 3 Arabic-specific adaption stages: extra pretraining on 10B Arabic tokens, supervised fine-tuning using Arabic-labeled data, and Reinforcement Learning with Human Feedback~\cite{ouyang2022training} tuning, before being fine-tuned on each task of ALUE separately.

\subsection{Discussion}
\label{sec:Discussion}

Despite their multilingual capabilities, some recent studies~\cite{chatgpt2022,openai2023gpt4} have shown that the most prominent LLM perform poorly on tasks involving the Arabic language. For example, \citet{khondaker2023gptaraeval} estimated\footnote{on an excerpt of 200 examples from each ORCA task.} a gap of 17.5\% on ORCA score between chatGPT~\cite{chatgpt2022} (10-shot) and \marbert{-v2}~\cite{abdul2021arbert}. It is also noteworthy that the latter model already underperforms \jaberv{} by 4.1\% on the ORCA leaderboard.

In addition, the performance of JASMINE~\cite{nagoudi2023jasmine}, a monolingual Arabic decoder-only LLM pre-trained on a dataset roughly equivalent to that of \arbert{v2}, reveals some shortcomings in handling Arabic Natural Language Understanding (NLU) tasks. For instance, in the best configuration, JASMINE (6.7B parameters with 24 in-context examples) scores 5.8\% on the dialect-R ORCA task, compared to 67.8\% for \jaberv{}; 57.3\% on the sarcasm task, compared to 74.3\% for \jaberv{}; and 45.5\% on the sentiment task, compared to 80.2\% for \jaberv{}, respectively.

These observations strongly suggest that the near future should witness multiple efforts to develop robust Arabic LLMs that can perform in Arabic as effectively as their counterparts do in English. Our work lays the groundwork in this direction, as we firmly believe that unlocking the full potential of LLMs for Arabic will require a mega-scale high-quality pretraining dataset.

\section{Conclusion and Future Work}
\label{sec:Conclusion}

In this work, we demonstrated the significance of scaling up pretraining data in enhancing the performance of Arabic PLMs, outperforming factors like model size or architecture. We developed four new Arabic PLMs (\jaberl, \jaberv, \atsv) with strong results on ALUE and ORCA leaderboards. Future work will focus on creating an Arabic LLM and addressing language-specific challenges in its development.

\section*{Limitations}

The main limitation of our work is the lack of measurement for improvements in Arabic generative tasks. This is primarily due to the absence of a well-established benchmark for these tasks, comparable in caliber to ALUE and ORCA. Although the paper on Dolphin~\cite{nagoudi2023dolphin}, which introduces an Arabic natural language generation benchmark with 50 tasks, was recently published, neither the leaderboard nor the datasets were available at the time of writing our paper. Additionally, our study has a limitation in that it does not include experiments with decoder-only models, as discussed in ~\cite{radford2018improving_gpt}.


\section*{Acknowledgements}
We thank Mindspore\footnote{\url{https://www.mindspore.cn/}} for the partial support of this work, which is a new deep learning computing framework.

\normalem 
\bibliography{custom}
\bibliographystyle{acl_natbib}

\clearpage
\newpage
\appendix

\section{Pretraining Details}
\label{sec:Pretraining Details}

\subsection{Data}
\label{sec:Pretraining Data}

Our pretraining data primarily derives from the dataset used by~\cite{alghamdi2023aramus}, which is recognized as the largest reported Arabic pretraining corpus to date. The data was mined from 90 shards of monthly Common Crawl, covering the period from 2013 to 2022. It encompasses 8.7TB of plain Arabic text, which was reduced to 512GB after de-duplication and extensive data filtering, following the procedure described in ~\cite{ghaddar2022revisiting}. This pretraining corpus is 4 times larger than the one used to pretrain \jaber{}, \ats{}, and \atb{} models of \cite{ghaddar2022revisiting}, and twice larger than the one used to pretrain \arabert{v2}~\cite{elmadany2023orca} and JASMINE~\cite{nagoudi2023jasmine}.

\subsection{Models and Implementation Details}

We pretrained four new models from scratch, each derived from the \jaber{}, \ats{}, \atb{} models developed in ~\cite{ghaddar2022revisiting}:

\begin{itemize}
    \item \textbf{\jaberv{}} is a 12-layer Arabic \bert{-base} that is equivalent to \jaber{}.
    \item \textbf{\jaberl{}} is a 6-layer version of \jaberv{}.
    \item \textbf{\atsv{}} and \textbf{\atbv{}} is an Arabic T5-small and T5-base models that follow \ats{} and \atb{}, respectively.
\end{itemize}

Following~\cite{ghaddar2022revisiting}, we use a byte-level byte pair encoding (BBPE)~\cite{wei2021training} tokenizer, to process sub-tokens and learn a new vocabulary of size 64k.
We perform pretraining on 8 nodes each with 8 NVIDIA Tesla V100 GPUs with 32GB of memory. For encoder-only models, we set the initial learning rate to 1e-4, with 10000 warm-up steps, and used AdamW ~\cite{loshchilov2017decoupled} optimizer with a learning rate linear decay. We train for a total of 15 epochs with a maximum sequence length of 128 tokens, while setting the per GPU batch size to 64. For encoder-decoder models, we use the Adafactor optimizer~\cite{adafactor2018} with an initial learning rate of 1 and inverse square-root decay continuing until the end of pre-training. The maximum sequence length is set to 512 for the encoder and 114 for the decoder, and we train for 200k steps. Additionally, the per-GPU batch size is set to 56 for \ats{} and 16 for \atb{}, respectively.

\section{Fine-tuning Details}
\label{sec:Fine-tuning Details}

\subsection{Datasets}

We experiment with 2 well-established Arabic standards benchmarks ALUE~\cite{seelawi2021alue} and ORCA~\cite{elmadany2023orca}. Both benchmarks feature official train/dev/test splits, have a public leaderboard, and include privately held test sets. ALUE comprises 8 NLP tasks, 6 of which focus on Arabic dialects. Additionally, the training examples in ALUE are generally characterized by a relatively small number of training examples and a short sequence length. We report Pearson
correlation on SVREG, accuracy on XNLI, Jaccard similarity on SEC, and F1 score for the other tasks. ALUE was built on top of the following datasets: FID~\cite{ghanem2019idat}, MDD~\cite{bouamor-etal-2019-madar}, MQ2Q~\cite{seelawi-etal-2019-nsurl}, OOLD and OHSD~\cite{mubarak-etal-2020-overview}, SVREG and SEC~\cite{mohammad-etal-2018-semeval}, XNLI~\cite{conneau2018xnli}.

On the other hand, ORCA is a considerably larger benchmark, encompassing 29 tasks. These tasks cover a wide range of areas, including sequence classification, sequence pair classification, regression, part of speech tagging, named entity recognition, topic classification, word sense classification, and question answering. The evaluation metric is the macro F1 score for all tasks, except for sts-r, which uses Spearman correlation. The datasets to build ORCA were crafted from the following individual datasets: ans-st~\cite{khouja-2020-stance}, baly-st~\cite{baly-etal-2018-integrating}, xnli~\cite{conneau2018xnli}, qa~\cite{abdul2021arbert}, sts-r~\cite{mohammad-etal-2018-semeval}, sts-c~\cite{seelawi-etal-2019-nsurl}, abusive~\cite{mulki-etal-2019-l}, adult~\cite{mubarak-etal-2021-adult}, age~\cite{abdul-mageed-etal-2020-aranet}, claim~\cite{khouja-2020-stance}, dangerous~\cite{alshehri-etal-2020-understanding}, dialect-\{B,R,C\}~\cite{abu-farha-magdy-2020-arabic,zaidan-callison-burch-2014-arabic,bouamor-etal-2019-madar,abdelali-etal-2021-qadi,el-haj-2020-habibi}, emotion~\cite{abdul-mageed-etal-2020-aranet}, gender~\cite{abdul-mageed-etal-2020-aranet}, hate~\cite{mubarak-etal-2020-overview}, irony~\cite{ghanem2019idat}, machine G.~\cite{nagoudi-etal-2020-machine}, offensive~\cite{mubarak-etal-2020-overview}, sarcasm~\cite{abu-farha-magdy-2020-arabic}, sentiment~\cite{abdul-mageed-etal-2020-aranet}, sedar~\cite{ghaddar2020sedar}, ner-aqmar~\cite{schneider2012coarse}, ner-anerc~\cite{benajiba2007anersys}, pos-dia~\cite{darwish2018multi}, pos-xglue~\cite{liang2020xglue}, topic~\cite{abbas2011evaluation}, wsd~\cite{app11062567}. We direct readers to \cite{seelawi2021alue} and \cite{elmadany2023orca} for more details on the ALUE and ORCA benchmarks, respectively.

\subsection{Baselines}

Table~\ref{tab:model_config_data} shows the architecture and the size of the pretraining data for all baseline models included in this study. This includes the models under the \bom{} configuration presented in Tables~\ref{tab:orca_test} and ~\ref{tab:orca_dev}.

\begin{table}[!ht]
\centering
\small
\resizebox{\columnwidth}{!}{
\begin{tabular}{l|c|c|c}
\toprule
\bf Model Name & \bf Config. & \bf PDS  \\
\midrule

\arabert{}~\cite{antoun2020arabert} & \bert{-base} &  27GB \\
AraElectra~\cite{antoun-etal-2021-araelectra} & Electra-base &  77GB \\
ArabicBERT~\cite{safaya2020kuisail} & \bert{-base} &  95GB \\
Arabic-ALBERT~\cite{safaya2020kuisail} & Albert-base &  33GB \\
QARiB~\cite{chowdhury2020improving} & \bert{-base} &  97GB \\
CAMeLBERT~\cite{inoue2021interplay} & \bert{-base} &  167GB \\
\arbert{}\footnotesize{~\cite{abdul2021arbert}} & \bert{-base} & 61GB\\
\marbert{}\footnotesize{~\cite{abdul2021arbert}} & \bert{-base} &  128GB \\
\marbert{v2}\footnotesize{~\cite{abdul2021arbert}} & \bert{-base} &  198GB \\
\arbert{v2}\footnotesize{~\cite{elmadany2023orca}} & \bert{-base} &  243GB \\

\jaber{}~\cite{ghaddar2021jaber} & \bert{-base} & 128GB \\
\saber{}~\cite{ghaddar2021jaber} & \bert{-large} & 128GB \\
\ats{}~\cite{ghaddar2022revisiting} & T5-small & 128GB \\ 
\atb{}~\cite{ghaddar2022revisiting} & T5-base & 128GB \\
\artb{}~\cite{nagoudi2022_arat5} & T5-base & 248GB \\ 
JASMINE~\cite{nagoudi2022_arat5} & GPT & 248GB \\ 
\aramus{}~\cite{alghamdi2023aramus} & T5-xxl &  529GB \\
\bottomrule
\end{tabular}
}
\caption{Names of the baseline models included in this study, along with their configurations (Configs.)  and the sizes of their pretraining data (PDS).}
\label{tab:model_config_data}
\end{table}

\subsection{Implementation Details}

\begin{figure*}[!htb]
    \centering
    \includegraphics[width=15cm]{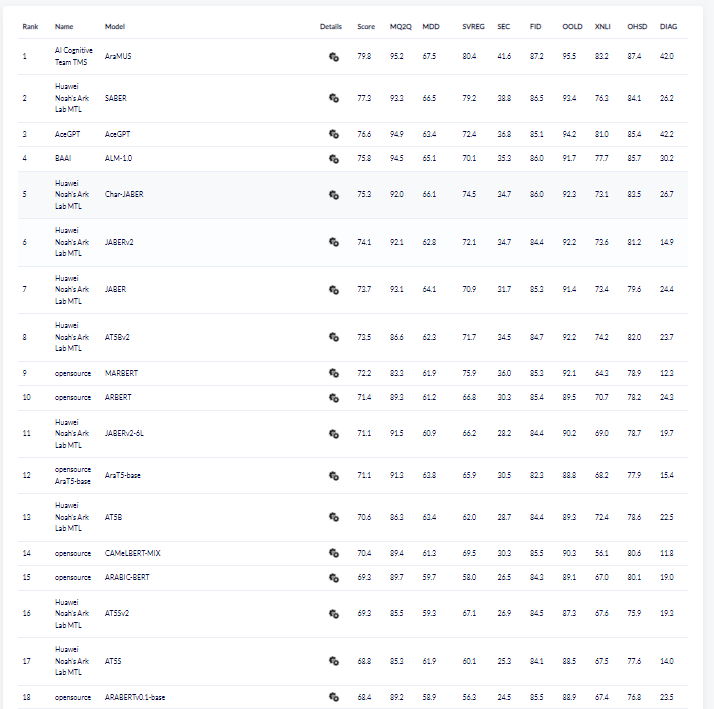}
    \caption{Screenshot of ALUE leaderboard as of 01/01/2024. We turn off our submission private during the anonymity period of the submission.}
    \label{fig:alue_leaderboard}
\end{figure*}

\begin{figure*}[!htb]
    \centering
    \includegraphics[width=15cm]{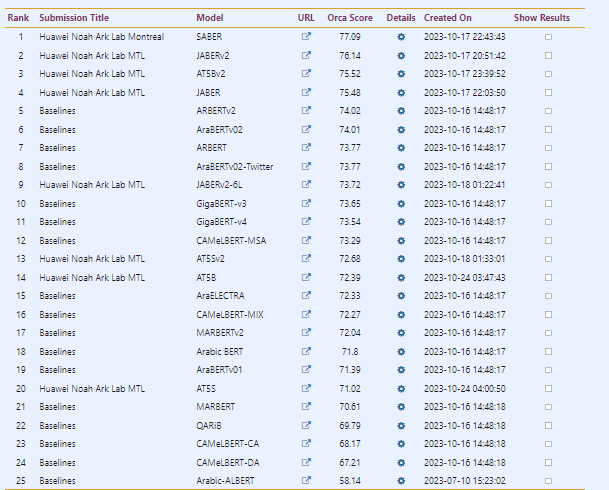}
    \caption{Screenshot of ORCA leaderboard as of 01/01/2024. We turn off our submission private during the anonymity period of the submission.}
    \label{fig:orca_leaderboard}
\end{figure*}

We conducted extensive experiments to ensure a fair comparison of performances between models following ~\cite{ghaddar2022revisiting}. This process involved conducting a grid search to identify the best hyperparameters, which were selected based on performance on the development set. Subsequently, we carried out five runs using different random seeds and reported the means and standard deviation to ensure the statistical significance of the development set results. The checkpoint from the best-performing run was then used to submit results to the leaderboard's test set. We applied this procedure consistently across each model and for every task. Figures~\ref{fig:alue_leaderboard} and ~\ref{fig:orca_leaderboard} present anonymized screenshots showing the ranks of our models on the ALUE and ORCA leaderboards, respectively.

For all encoder-only models, we use AdamW optimizer~\cite{kingma2014adam} with learning rate with linear decay. We search the learning rate from \{7e-6, 2e-5, 5e-5\}. For all encoder-decoder models, we use the Adafactor~\cite{adafactor2018} with inverse square root decay and pick a learning rate from \{1e-3, 1e-4, 1e-5\}. For all models, we searched batch size from \{8, 16, 32, 64, 128\}, hidden dropout from \{0.1, 0.2, 0.3, 0.4\}, and fixed the epoch number to 30. Our fine-tuning code utilizes the PyTorch version~\cite{paszke2019pytorch} of the HuggingFace Transformers library~\cite{wolf2020transformers}. All experiments were conducted on a single NVIDIA Tesla V100 GPU.

\section{Dev Set Results}
\label{sec:Dev Set Results}

\begin{table*}[!htp]
\small
\centering
\resizebox{\textwidth}{!}{
\begin{tabular}{l|c|cccccccc|c}
\toprule
\textbf{Model} & \textbf{\#Params} & \textbf{MQ2Q}& \textbf{MDD}& \textbf{SVREG}& \textbf{SEC}& \textbf{FID}& \textbf{OOLD}& \textbf{XNLI}& \textbf{OHSD} & \textbf{Avg.}\\

\midrule
\multicolumn{10}{c}{\textit{T5 Models}}\\
\midrule

\ats* & 109M & 73.0$\pm$0.1 & 63.1$\pm$0.3 & 75.6$\pm$1.6 & 41.3$\pm$0.5 & 82.1$\pm$0.6 & 88.4$\pm$0.2 & 67.9$\pm$0.3 & 81.0$\pm$1.8 & 71.5$\pm$0.7   \\
\atsv & 109M & 69.2$\pm$0.6 & 61.5$\pm$0.2 & 85.4$\pm$0.4 & 43.0$\pm$0.4 & 83.7$\pm$0.6 & 90.4$\pm$0.4 & 67.4$\pm$0.1 & 80.1$\pm$1.5 & 72.6$\pm$0.5   \\

\artb* & 289M  & 70.5$\pm$2.1 & 63.6$\pm$0.2     & 80.8$\pm$1.3 & 44.0$\pm$0.6 & 82.3$\pm$0.4     & 90.5$\pm$0.4 & 72.5$\pm$1.5 & 78.3$\pm$1.4     & 73.0$\pm$1.0  \\
\atb*  & 296M     & \underline{73.7$\pm$0.1} & \underline{64.7$\pm$0.2} & 78.1$\pm$2.4     & 43.8$\pm$0.7     &  83.1$\pm$0.5 & 90.0$\pm$0.4     & 72.2$\pm$0.4     & 81.2$\pm$2.1 & 73.3$\pm$0.9 \\
\atbv & 296M & 69.7$\pm$0.5 & 62.9$\pm$0.2 & \underline{\bf 88.2$\pm$0.5} & \underline{46.7$\pm$0.9} & \underline{84.2$\pm$0.3} & \underline{\bf 92.5$\pm$0.4} & \underline{\bf 73.9$\pm$0.4} & \underline{83.2$\pm$0.9} & \underline{75.2$\pm$0.5}   \\

\midrule
\multicolumn{10}{c}{\textit{\bert{} Models}}\\
\midrule
\jaberl & 86M & 74.0$\pm$0.5 & 62.2$\pm$0.4 & 86.4$\pm$0.4 & 45.6$\pm$0.6 & 84.0$\pm$0.4 & 91.3$\pm$0.3 & 68.3$\pm$0.5 & 83.4$\pm$0.9 & 74.4$\pm$0.5   \\
\jaber* & 135M & 75.1$\pm$0.3 & \underline{\bf 65.7$\pm$0.3} & \underline{87.4$\pm$0.7} & 46.8$\pm$0.8 & \underline{\bf 84.8$\pm$0.3} & \underline{92.2$\pm$0.5} & 72.4$\pm$0.7 & 85.0$\pm$1.6 & 76.2$\pm$0.7 \\
\jaberv & 135M  & \underline{\bf 75.2$\pm$0.5} & 64.4$\pm$0.2 & \underline{87.4$\pm$0.5} & \underline{\bf 47.8$\pm$0.9} & 84.2$\pm$0.3 & \underline{92.2$\pm$0.4} & \underline{73.2$\pm$0.4} & \underline{\bf 88.1$\pm$0.9} & \underline{\bf 76.6$\pm$0.5} \\

\bottomrule

\end{tabular}
}
\caption{\textsc{Dev} set performances and standard deviations over 5 runs on the ALUE benchmark of T5 and \bert{} models. Entries in bold indicate the best results across all models, while entries that are underlined denote the best results within each respective category. * indicate results copied directly from ~\cite{ghaddar2022revisiting}}
\label{tab:dev_alue}
\end{table*}

\begin{table*}[ht] 

\centering
 \renewcommand{\arraystretch}{1.25}
\resizebox{1\textwidth}{!}{%
\begin{tabular}{l@{\hspace{0.75\tabcolsep}}l@{\hspace{0.75\tabcolsep}}|c@{\hspace{0.75\tabcolsep}}c@{\hspace{0.75\tabcolsep}}|c@{\hspace{0.75\tabcolsep}}c@{\hspace{0.75\tabcolsep}}c@{\hspace{0.75\tabcolsep}}c@{\hspace{0.75\tabcolsep}}c@{\hspace{0.75\tabcolsep}}c@{\hspace{0.75\tabcolsep}}c@{\hspace{0.75\tabcolsep}}c@{\hspace{0.75\tabcolsep}}c}  

 \toprule  
\textbf{\colorbox{blue!0}{Cluster}}  &
\textbf{\colorbox{blue!0} {Task}}&
\textbf{\colorbox{purple!20}{{\arbert}}}  &
\textbf{\colorbox{purple!20}{{\bom}}}&
\textbf{\colorbox{yellow!30}{{\saber}}} &
\textbf{\colorbox{purple!20}{{\jaberv  }}}&
\textbf{\colorbox{green!20}{{\atbv }}} &
\textbf{\colorbox{purple!20}{{\jaber }}} &
\textbf{\colorbox{gray!20}{{\jaberl }}}  & 
\textbf{\colorbox{blue!20}{{\atsv  }}} &
\textbf{\colorbox{green!20}{{\atb }}} &
\textbf{\colorbox{blue!20}{{\ats }}}  
\\
\toprule 

\multicolumn{1}{l}{\multirow{11}{*}{\textbf{SC}}} 

& abusive   &  76.0 & 79.6 & \textbf{81.8$\pm$1.1}  & \underline{80.2$\pm$0.6}       & 79.0$\pm$0.8     & 80.1$\pm$1.5 & 79.3$\pm$1.1         & 78.1$\pm$0.7      & 75.6$\pm$1.3 & 76.7$\pm$0.8 \\
& adult & 89.7 & 91.0  & \textbf{92.6$\pm$0.1}  & \underline{92.5$\pm$0.1}  & 91.5$\pm$0.3     & 91.7$\pm$0.1 & 91.8$\pm$0.3         & 91.3$\pm$0.4      & 91.2$\pm$0.2 & 90.9$\pm$0.3 \\
& age  &  45.6 & 47.3    & 99.7$\pm$0.0  & 99.8$\pm$0.0       & \underline{\textbf{99.9$\pm$0.0}}     & 99.7$\pm$0.0 & 99.9$\pm$0.0         & 99.9$\pm$0.0      & 99.7$\pm$0.0 & 99.7$\pm$0.0 \\
&  claim &  67.4 & 71.5 & \underline{\textbf{72.4$\pm$1.1}}  & 70.6$\pm$0.9       & 70.4$\pm$1.1     & 71.6$\pm$0.4 & 67.5$\pm$1.1         & 67.8$\pm$1.1      & 68.7$\pm$0.8 & 66.6$\pm$1.0 \\
 & dangerous &  65.0 & 67.3  &  66.7$\pm$13.1 & 76.2$\pm$0.7       & \underline{\textbf{76.4$\pm$0.6}}     & 70.4$\pm$1.4 & 75.1$\pm$0.8         & 73.5$\pm$0.9      & 67.7$\pm$0.6 & 67.7$\pm$0.7 \\
& dialect-B &   86.9 & 87.6  & \textbf{90.1$\pm$0.3}  & 89.4$\pm$0.3       & 89.5$\pm$0.1     & \underline{89.9$\pm$0.2} & 89.5$\pm$0.2         & 89.0$\pm$0.3      & 89.3$\pm$0.3 & 89.2$\pm$0.1 \\			
&  dialect-R  &   65.2 & 69.2  & \textbf{92.5$\pm$0.1}  & \underline{91.2$\pm$0.2}       & 91.0$\pm$0.1     & 90.5$\pm$0.3 & 89.7$\pm$0.4         & 89.4$\pm$0.4      & 89.5$\pm$0.2 & 88.8$\pm$0.2 \\
&  dialect-C &  35.7 & \underline{36.5}  & \underline{38.5$\pm$0.6}  & 36.4$\pm$0.2       & 34.5$\pm$0.5     & 35.9$\pm$0.4 & 33.5$\pm$0.4         & 31.3$\pm$0.5      & 32.5$\pm$0.6 & 29.8$\pm$0.3 \\
&  emotion-cls &   64.8 & \underline{\textbf{70.8}} & 70.2$\pm$1.2  & 67.8$\pm$0.7       & 65.7$\pm$1.0     & 65.2$\pm$0.5 & 64.6$\pm$0.9         & 62.6$\pm$0.5      & 61.6$\pm$0.5 & 59.9$\pm$0.7 \\
& emotion-reg  &  67.7 & \underline{\textbf{74.3}} & 74.0$\pm$1.3  & 73.8$\pm$0.7       & 72.5$\pm$0.1     & 69.5$\pm$0.9 & 69.8$\pm$0.4         & 62.8$\pm$1.4      & 62.6$\pm$0.9 & 57.1$\pm$0.8 \\
& gender  &  63.2 & \underline{\textbf{67.6}} & 65.9$\pm$0.5  & 65.6$\pm$0.3       & 65.5$\pm$0.2     & 63.9$\pm$0.3 & 65.1$\pm$0.3         & 64.4$\pm$0.1      & 63.3$\pm$0.3 & 62.3$\pm$0.2 \\

 &  hate   & 82.3 & 85.3 & \textbf{87.8$\pm$1.4}  & \underline{85.7$\pm$0.9}       & 82.8$\pm$1.4     & 85.2$\pm$0.2 & 81.7$\pm$0.4         & 79.1$\pm$0.5      & 79.7$\pm$0.6 & 78.6$\pm$0.5 \\
  &  irony  &  83.8 & \underline{\textbf{85.2}}  & 85.0$\pm$0.6  & 84.9$\pm$0.8       & 84.6$\pm$0.9     & 84.4$\pm$0.3 & 84.6$\pm$1.1         & 84.9$\pm$0.4      & 82.6$\pm$1.2 & 82.4$\pm$0.8 \\
  &  offensive &  89.6 & 92.2 &  \textbf{92.8$\pm$0.2}  & 91.9$\pm$0.2       & \underline{92.5$\pm$0.3}     & 91.7$\pm$0.3 & 91.3$\pm$0.1         & 90.5$\pm$0.6      & 89.7$\pm$0.7 & 88.6$\pm$0.4 \\

  &  machine G. &  87.9 & \underline{\textbf{90.7}}  &  90.0$\pm$0.3  & 88.7$\pm$0.4       & 87.7$\pm$0.3     & 89.2$\pm$0.4 & 86.9$\pm$0.2         & 86.2$\pm$0.4      & 88.0$\pm$0.4 & 86.0$\pm$0.6 \\
  &  sarcasm &  74.2 & \underline{\textbf{79.6}} & 76.8$\pm$0.7  & 76.4$\pm$0.6       & 75.2$\pm$0.2     & 76.9$\pm$0.5 & 75.9$\pm$0.2         & 77.1$\pm$0.4      & 75.5$\pm$0.6 & 71.7$\pm$1.2 \\

& sentiment  &   78.6 & 80.9 & \textbf{89.6$\pm$0.3}  & 88.6$\pm$0.1       & \underline{89.0$\pm$0.2}     & 88.8$\pm$0.4 & 88.5$\pm$0.3         & 87.4$\pm$0.2      & 88.3$\pm$0.4 & 87.4$\pm$0.5 \\ 


\multicolumn{1}{l}{\multirow{3}{*}{\textbf{SP}}} 
 &  ner-anerc.   & 90.8 & \underline{\textbf{90.9}} & 90.5$\pm$0.2  & 89.9$\pm$0.2       & 89.9$\pm$0.3     & 90.1$\pm$0.3 & 89.5$\pm$0.3         & 89.7$\pm$0.2      & 90.3$\pm$0.3 & 89.2$\pm$0.3 \\
  &  ner-aqmar   &   81.7 & 81.7 & 81.4$\pm$0.3  & 81.5$\pm$0.6       & \underline{\textbf{82.4$\pm$0.1}}     & 81.4$\pm$0.2 & 81.0$\pm$0.4         & 80.3$\pm$0.5      & 81.2$\pm$0.4 & 81.7$\pm$0.4 \\
  & pos-dia  &   93.9 & 94.7  & 95.9$\pm$0.1  & \underline{\textbf{95.9$\pm$0.1}}       & 95.5$\pm$0.1     & 95.3$\pm$0.1 & 95.3$\pm$0.1         & 94.9$\pm$0.2      & 94.6$\pm$0.1 & 94.4$\pm$0.1 \\ 
   &   pos-xglue &  52.6 &  \underline{\textbf{63.9}} & 49.5$\pm$4.0  & 61.3$\pm$0.6       & 61.7$\pm$0.6     & 41.6$\pm$1.0 & 55.1$\pm$0.4         & 53.9$\pm$1.1      & 42.1$\pm$2.0 & 36.8$\pm$0.6 \\

 \multicolumn{1}{l}{\multirow{3}{*}{\textbf{NLI}}} 

 &  ans-st &  91.0 & 93.3    & 95.2$\pm$0.4  & \underline{\textbf{95.2$\pm$0.4}}       & 92.4$\pm$1.8     & 93.7$\pm$0.5 & 88.8$\pm$0.7         & 87.4$\pm$0.9      & 87.5$\pm$1.7 & 86.0$\pm$2.4 \\
 &  baly-st   &    49.3 & 51.2    & 56.4$\pm$1.8  & 58.2$\pm$2.0       & \underline{\textbf{60.0$\pm$1.5}}     & 59.4$\pm$0.8 & 55.9$\pm$1.1         & 56.4$\pm$2.1      & 58.7$\pm$1.0 & 55.8$\pm$2.1 \\ 
  &  xlni  &  68.2 & 70.2   & 71.2$\pm$0.7  & \underline{\textbf{71.4$\pm$0.2}}       & 69.5$\pm$0.4     & 69.1$\pm$0.8 & 65.8$\pm$1.6         & 63.5$\pm$0.5      & 67.4$\pm$0.6 & 63.0$\pm$0.4 \\ 
  
\multirow{2}{*}{\textbf{STS}}  

  &  sts-r  & 71.9 & 76.0    & \textbf{86.0$\pm$0.6}  & \underline{85.8$\pm$0.5}       & 83.4$\pm$0.3     & 84.0$\pm$0.3 & 83.1$\pm$0.9         & 81.7$\pm$0.8      & 83.0$\pm$0.5 & 79.4$\pm$0.3 \\
  &  sts-c & 96.7 & \underline{\textbf{97.1}}   & 96.6$\pm$0.2  & 96.3$\pm$0.2       & 96.3$\pm$0.2     & 96.5$\pm$0.2 & 95.9$\pm$0.2         & 95.2$\pm$0.3      & 96.1$\pm$0.3 & 95.7$\pm$0.2 \\\hline 
     
    \textbf{TC} &  
  topic   &  94.0 &  94.6  & 94.6$\pm$0.1  & 94.3$\pm$0.3       & 94.5$\pm$0.2     & 94.4$\pm$0.1 & 94.1$\pm$0.2         & 93.9$\pm$0.2      & \underline{\textbf{94.7$\pm$0.1}} & 94.2$\pm$0.3 \\\hline

 \multirow{1}{*}{\textbf{QA}}  
   &  arlue-qa  & 61.5 & \underline{\textbf{61.6}}     & 59.3$\pm$0.7  & 53.3$\pm$0.0       & 60.0$\pm$0.6     & 59.7$\pm$0.5 & 54.1$\pm$0.6         & 53.4$\pm$0.4      & 57.2$\pm$0.0 & 52.8$\pm$0.4 \\ \hline
 
\textbf{WSD} &
  ar-wsd   & \textit{?} & \textit{?}    & \textbf{76.6$\pm$0.3}  & 74.5$\pm$0.4       & 73.7$\pm$0.7     & \underline{75.1$\pm$0.7} & 71.8$\pm$0.6         & 71.3$\pm$0.7      & 74.1$\pm$0.7 & 71.3$\pm$0.7 \\ \hline
							
  \toprule  
  
\multicolumn{2}{l}{\textbf{ORCA score}} &  74.4 & - & \textbf{80.0$\pm$1.1}  & \underline{79.9$\pm$0.5}       & 79.6$\pm$0.5     & 78.8$\pm$0.5 & 78.1$\pm$0.5         & 77.2$\pm$0.5      & 77.0$\pm$0.6 & 75.3$\pm$0.5 \\ \hline 			
\toprule  

\end{tabular}}

\caption{\textsc{Dev} set performances and standard deviations over 5 runs on the ORCA benchmark. The format and notation of this table are consistent with those in Table~\ref{tab:orca_test}. Entries marked with a \textit{?} indicate results that could not be retrieved from the ~\cite{elmadany-etal-2023-orca} paper. The table formatting follows the style used in~\cite{elmadany-etal-2023-orca}.}
\label{tab:orca_dev}.

\end{table*}

\clearpage

\end{document}